\documentclass[letterpaper]{article} 
\usepackage[preprint]{aaai24}  
\usepackage{times}  
\usepackage{helvet}  
\usepackage{courier}  
\usepackage[hyphens]{url}  
\usepackage{graphicx} 
\usepackage{natbib}  
\usepackage{caption} 
\usepackage{algorithm}
\usepackage{algorithmic}

\usepackage[hyphens]{url}
\usepackage{minitoc}
\usepackage{natbib}
\usepackage{subcaption}
\usepackage{dsfont} 
\usepackage{amssymb}
\usepackage{amsthm}
\usepackage{stmaryrd} 
\usepackage{times}
\usepackage{epsfig}
\usepackage{graphicx}
\usepackage{amsmath}
\usepackage{dsfont}
\usepackage{amssymb}
\usepackage{booktabs}
\usepackage{multirow}
\usepackage{pifont}
\usepackage{comment}
\usepackage{caption}
\usepackage{tikz}
\usepackage{comment}
\usepackage{amsmath,amssymb} 
\usepackage{color}
\definecolor{Gray}{gray}{0.85}
\usepackage{array}
\usepackage{eqparbox}

\newcommand{\round}[1]{\left\lfloor #1 \right\rceil}
\newcommand{\floor}[1]{\left\lfloor #1 \right\rfloor}
\newcommand{\R}[1]{\mathbb{R}^{#1}}

\usepackage{newfloat}
\usepackage{listings}
\DeclareCaptionStyle{ruled}{labelfont=normalfont,labelsep=colon,strut=off} 
\floatstyle{ruled}
\newfloat{listing}{tb}{lst}{}
\floatname{listing}{Listing}
\newcommand\blfootnote[1]{%
  \begingroup
  \renewcommand\thefootnote{}\footnote{#1}%
  \addtocounter{footnote}{-1}%
  \endgroup
}

\title{Gradient-Based Post-Training Quantization: Challenging the Status Quo}
\author{
    Edouard Yvinec$^{1,2}$ , Arnaud Dapogny$^2$ , Kevin Bailly$^{1,2}$
}

\affiliations{
    Sorbonne Université$^1$, CNRS, ISIR, f-75005, 4 Place Jussieu 75005 Paris, France \\
  Datakalab$^2$, 114 boulevard Malesherbes, 75017 Paris, France \\
  \texttt{ey@datakalab.com} \\
\blfootnote{pre-print}
}

\usepackage{bibentry}

\begin{document}

\maketitle

\begin{abstract}
Quantization has become a crucial step for the efficient deployment of deep neural networks, where floating point operations are converted to simpler fixed point operations. In its most naive form, it simply consists in a combination of scaling and rounding transformations, leading to either a limited compression rate or a significant accuracy drop. Recently, Gradient-based post-training quantization (GPTQ) methods appears to be constitute a suitable trade-off between such simple methods and more powerful, yet expensive Quantization-Aware Training (QAT) approaches, particularly when attempting to quantize LLMs, where scalability of the quantization process is of paramount importance. GPTQ essentially consists in learning the rounding operation using a small calibration set. In this work, we challenge common choices in GPTQ methods. In particular, we show that the process is, to a certain extent, robust to a number of variables (weight selection, feature augmentation, choice of calibration set). More importantly, we derive a number of best practices for designing more efficient and scalable GPTQ methods, regarding the problem formulation (loss, degrees of freedom, use of non-uniform quantization schemes) or optimization process (choice of variable and optimizer). Lastly, we propose a novel importance-based mixed-precision technique. Those guidelines lead to significant performance improvements on all the tested state-of-the-art GPTQ methods and networks (e.g. +6.819 points on ViT for 4-bit quantization), paving the way for the design of scalable, yet effective quantization methods.
\end{abstract}

\section{Introduction}
From the early emergence of deep neural networks \citep{lecun1989backpropagation} to the most recent breakthroughs \citep{zhang2022opt}, the deep learning community observed two constant trends: the increase in predictive accuracy and the exponential growth of neural network size. Concurrently, deep neural networks (DNNs) acceleration techniques have been introduced in order to address the latter which lead to the confinement of DNNs deployment on large devices. These methods are usually divided into five categories: pruning \citep{frankle2018lottery}, tensor decomposition \citep{yu2017compressing}, knowledge distillation \citep{hinton2015distilling}, efficient model design (e.g. using NAS techniques \citep{elsken2018neural}) and quantization \citep{gholami2021survey}.

Among the aforementioned methods, quantization appears to offer the best compression results for any given deep architecture \citep{guo2018survey}. 
It consists in the conversion of the format of the computations performed by the model. In practice, most models are trained using either 32 or 16 bits floating point operations \citep{guo2018survey}. At inference time, lower latency can be achieved through the use of lower bitwidth representations and fixed-point computations \citep{gholami2021survey}. In its most effective form, quantization is performed during the training process and leads to the most remarkable compression rates. The core challenge comes from the rounding operation from the conversion to fixed point values. This rounding transformation introduces zero gradients almost everywhere. Consequently, quantization aware training (QAT) aims at designing workarounds. As a result, we observe two categories of QAT methods: straight through estimation (STE) \cite{esser2019learned,bhalgat2020lsq+} and soft approximation \cite{gong2019differentiable}. While the former proposes to replace the zero gradient function of the rounding operation, the latter approximates the rounding function by a non-zero derivative transformation. All the aforementioned methods suffer from the same pitfall: they introduce a significant overhead during training, which reaches its maximum for binary neural networks \cite{zhang2022pokebnn} where the process takes up to 10 times more time and compute.

In order to circumvent this limitation, \citet{nagel2019data} introduced a novel quantization paradigm: data-free quantization. In this set-up, the authors leverage the vast number of pre-trained models available off the shelf. The model is then quantized based on its trained weights and statistics stored in the batch-normalization layers. With the growing concerns for data privacy as well as its low processing cost, data-free quantization sparked numerous subsequent research \cite{li2021mixmix,squant2022,yvinec2023power}. However, these methods still fall short in terms of compression rates as compared to QAT. Consequently, \citet{nagel2020up} proposed the first gradient-based post-training quantization (GPTQ), as a middle ground method. In general, GPTQ methods leverage a small calibration set in order to learn the rounding operation and improve data-free quantization performance \cite{nagel2020up,li2021brecq,wei2022qdrop,liu2023pd}. As a result, the quantized model usually achieve performances close to QAT models models in terms accuracy \textit{v.s.} speed trade offs. Thus, GPTQ methods may constitute a suitable middle ground between data-free and QAT methods, which may be well-suited for efficiently producing accurate quantized networks, especially considering LLMs , where memory bandwidth and training time are often bottleneck.

During the last years, GPTQ methods have been improved through different aspects. In its first implementation, GPTQ optimizes each layer individually and sequentially from first to last layer, using Adam \cite{kingma2014adam} optimizer. Many methods have been proposed and challenged some hypothesis made in AdaRound. For example, in BrecQ, \citet{li2021brecq} proposed to optimize computational blocks rather than a single layer at a time. Example of computational blocks are residual blocks for ResNet architectures \cite{he2016deep} or transformer blocks in transformers \cite{vaswani2017attention}. The resulting quantized model systematically outperformed AdaRound. Second, authors in QDrop \cite{wei2022qdrop} proposed stochastic application of the quantization process, an increase the numbers of optimization steps. Third, PD-Quant \cite{liu2023pd} introduced a novel loss term that enforces preservation of the batch norm statistics. However, this method fails to maintain several good properties of AdaRound: no need to infer the whole network during optimization, and no need for specific layers during training (PD-Quant needs batch-normalization layers). Furthermore, it has a limited impact on recent DNN architectures (e.g. transformers, that do not use batch normalization) and requires more computations, in order to optimize over the final predictions. Last but not least, other works investigating GPTQ focus on generating relevant training examples \cite{xu2020generative}. The core idea of these approaches consists in optimizing a white noise input using regularization tricks in order to generate calibration examples, akin to \cite{yin2020dreaming,li2021mixmix}. However, this data generation process is very computationally intensive.

In this work, we propose to question a number of hypothesises of GPTQ methods. Every point raised is motivated by both mathematical intuition and empirical evidence suggesting that further improvements can be achieved. First, we show that the GPTQ process is robust to a number of factors of variations, including:

\begin{itemize}
    \item Weight selection based upon ambiguity, as used e.g. in \cite{squant2022}, cannot be employed in a straightforward fashion in GPTQ.
    \item Selection based on weight magnitude \cite{yvinec2023power} also does not bring consistently better results.
    \item We show that feature augmentation (dropout, mixup, cutout, noise) does not significantly impact the performance.
    \item We show that the GPTQ process is, to a certain extent, robust to the choice of the calibration set. In particular, using adversarial or out-of-domain examples for calibration still allows for relatively efficient quantization.
\end{itemize}

More importantly, we show that it is sensitive to other factors, and hence we derice a number of guidelines:

\begin{itemize}
    \item Using losses other than $l_2$ does not bring consistent performance gains.
    \item Biases shall not be optimized due to the risk of overfitting on small calibration sets.
    \item We show that Adamax optimizer \cite{kingma2014adam} brings a steady improvement in every tested scenario.
    \item We propose adaptations of the GPTQ framework to handle non-uniform quantization methods (log or power quantization \cite{yvinec2023power}), which significantly increase the accuracy especially on low-bit floating point formats.
    \item we provide a novel importance-based mixed-precision approach for assigning the appropriate bit-width to each neuron across the whole network, and comes at practically no computationally cost.
\end{itemize}

The combination of these methods lead to significant, steady accuracy improvements on all the tested state-of-the-art GPTQ methods and architectures (e.g. +6.819 points on ViT for 4-bit quantization). As such, These results and guidelines shall pave the way for designing more efficient GPTQ baselines. In addition, it provides insights on the most relevant directions to explore: stemming from the AdaMAx optimizer specificities and further investigation on mixed-precision.

\section{Preliminaries}

In the following sections, we rely on empirical evidence to justify assumptions made in GPTQ optimization. We evaluate on ResNets \cite{he2016deep} as they constitute the reference benchmark on computer vision networks. Furthermore, we consider the more challenging (from a quantization viewpoint) MobileNet V2 \cite{sandler2018mobilenetv2}. Also, as transformers are becoming more and more mainstream in the machine learning landscape, is essential considering current trends. Consequently, we also consider ViT \cite{dosovitskiy2020image}. As quantization of networks trained on small datasets (e.g. Cifar-10) do not necessarily translate on real world use-cases, we benchmark these models on ImageNet classification \cite{imagenet_cvpr09}. We conduct our experiments using Torch \cite{torch} and models from the torchvision \cite{torchvision} package, on a single A100 gpu. Following AdaRound, we perform the optimization over 10K iterations, with batch size 32 over a calibration set of 1024 examples. Each results were obtained from 8 runs each.

Let's consider a trained neural network $F$ comprising $L$ layers ${(f_l)}_{l\in\{1,...,L\}}$ with weight tensors ${(W_l)}_{l\in\{1,...,L\}}$. GPTQ breaks down the network quantization into the optimization of computational groups $B_1,...,B_N$ where $B_i = f_l$ and $N=L$ for AdaRound and $B_i$ are bottleneck blocks or transformer blocks for BrecQ, \textit{i.e.} each layer is quantized separately, traditionally \textit{assigning the same precision (bit-width)} to all layers to the exception of the first and last ones. Let $X$ denote the raw input to block $B_n$ (\textit{\textit{i.e.} typically sampled from in-distribution data}, and \textit{without augmentation}). We want to minimize the reconstruction error, \textit{typically the Euclidean norm} between $B_n(X)$, the original full-precision block, and $B^Q_n(X)$ its quantized counterpart. Formally:
\begin{equation}
    \min_{\epsilon} \mathbb{E}_X\left[ \|B_n(X) - B^Q_n\|_2 \right].
\end{equation}
The $\epsilon$ variable encodes whether we are rounding up or down each scalar value individually, \textit{with the exception of weights}. Furthermore, each weight shall Formally, a layer is systematically quantized as follows
\begin{equation}
    f^Q_l (X) = \left\lceil\frac{X}{s_X}\right\rfloor s_X \times \left(\left\lfloor\frac{W_l}{s_{W_l}}\right\rfloor + \epsilon\right) s_{W_l}
\end{equation}
where $\lceil\cdot\rfloor$ and $\lfloor\cdot\rfloor$ correspond to the rounding and flooring operators respectively. \textit{Note that this formulation does not take into account the error introduced by quantization (weight ambiguity) nor the magnitude of the weight values.} The value of $\epsilon$ is initialized such that $\left(\left\lfloor\frac{W_l}{s_{W_l}}\right\rfloor + \epsilon\right) s_{W_l} = W_l$. This initialization method ensures that the optimization starts from a layer that is mathematically similar to the full-precision layer. Furthermore, constraining $\boldsymbol{\epsilon \in [0;1]}$ as in AdaRound ensures that the values stay close to the original values, limiting overfitting. \textit{This choice is however debatable, shall we consider e.g. non-uniform quantization schemes.})
The optimization is \textit{typically carried on using Adam optimizer \cite{kingma2014adam}}.

This choice of optimization leads to the following footprint cost: $\#X \times 32$ (batches of 32 elements) + $2\times \#W_l$ (for $W_l$ and $\epsilon$), where $\#A$ corresponds to the number of scalar elements in $A$.

In what follows, we question a number of methodological choices, as outlined in the introduction.

\section{GPTQ best practices}

\subsection{On the importance of the loss term}\label{sec:lossround}

As descripted above, GPTQ methods traditionnally minimize the euclidean distance between the full-precision and the quantized features. To assess the importance of this choice, we evaluate AdaRound and BrecQ with different choices of losses, \textit{i.e.} $l_1$, cosine dissimilarity, KL divergence, and $l_2$ loss.
Table \ref{tab:loss} shows comparisons between these losses. We observe that $l_2$ offers the most consistent overall performance. While the $l_1$ and KL losses may achieve higher results in certain cases, especially on ViT, they are generally less consistent across the tested architectures (in particular on ResNet 50) and quantization methods. 
\begin{table}[!t]
\caption{Influence of the similarity goal in GPTQ methods. We use the same quantization as in Table \ref{tab:ambiguity}. We highlight in bold the results that show an improvement over the baseline $l_2$.}
\label{tab:loss}
\centering
\setlength\tabcolsep{4pt}
\begin{tabular}{|c|c|c|c|c|}
\hline
Architecture & $l_1$ & cosine & KL & baseline: $l_2$ \\
\hline
\hline
\multicolumn{5}{|c|}{AdaRound} \\
\hline
\hline
ResNet 50 & 42.716 & 16.398 & 59.738 & 61.318 \\
MobileNet v2 & \textbf{65.630} & 0.100 & 0.100 & 65.314 \\
ViT b16 & \textbf{34.460} & 0.100 & \textbf{38.714} & 31.256 \\
\hline
\hline
\multicolumn{5}{|c|}{BrecQ}\\
\hline
\hline
ResNet 50 & 46.586 & 25.864 & 60.818 & 63.644 \\
MobileNet v2 & 49.932 & 0.100 & 0.100 & 50.130 \\
ViT b16 & 57.400 & 0.100 & \textbf{60.524} & 57.952 \\
\hline
\end{tabular}
\end{table}
As one of the core strength of GPTQ methods is their robustness to the task and architecture design, we conclude that \textit{the main loss term does significantly impact the performance of GPTQ methods but the current naive approach appears to be the more robust}.

\begin{table}[!t]
\caption{Performance of GPTQ methods in combination with intermediate features augmentation. We use the same quantization as in Table \ref{tab:ambiguity}. We highlight in bold the results that show an improvement over the baseline without augmentations.}
\label{tab:aug}
\centering
\setlength\tabcolsep{2pt}
\begin{tabular}{|c|c|c|c|c|}
\hline
Architecture & + Dropout & + Mix-up & + Cut-out & + Noise \\
\hline
\hline
\multicolumn{5}{|c|}{AdaRound} \\
\hline
\hline
ResNet 50    & \textbf{61.632} & 60.336 & 61.306 & 55.316 \\
MobileNet v2 & 65.178 & \textbf{65.788} & \textbf{65.864} & 62.506 \\
ViT b16         & 29.780 & 30.422 & 29.874 & \textbf{31.364} \\
\hline
\hline
\multicolumn{5}{|c|}{BrecQ}\\
\hline
\hline
ResNet 50    & \textbf{63.689} & 62.396 & \textbf{63.843} & 57.870 \\
MobileNet v2 & \textbf{50.321} & \textbf{50.151} & \textbf{51.052} & 46.974 \\
ViT b16         & 56.331 & 57.453 & 56.161 & 57.944 \\
\hline
\end{tabular}
\end{table}

\subsection{On the importance of data augmentation}\label{sec:auground}
Intermediate feature augmentations \cite{verma2019manifold} such as mix-up, cut-out or dropout are well known for their efficiency during full-precision training. Considering that our results show the good performance of quantization aware training inspired GPTQ methods such as NUPES \cite{yvinec2023nupes}, we propose to apply feature augmentation and measure its influence on GPTQ. This kind of augmentation can be formulated as finer-grained version of Q-Drop \cite{wei2022qdrop} which randomly drops the quantization of some layers during the optimization. By analogy, we rather randomly drop the quantization of some features.

In Table \ref{tab:aug}, we observe a similar behavior as for the loss function. While some data augmentation may lead to some improvements in specific cases, we do not observe stable results across all the benchmarked architectures. However it is worth noting that the best performing augmentations are dropout and cut-out which can be interpreted as sparse, local (layer-wise) versions of QDrop.

In summary, \textit{Feature augmentation does not significantly impact the performance of GPTQ methods} which we attribute to the fact, on the one's hand, feature reduces the risk of over-fitting which is important in GPTQ due to the small size of calibration sets. However, on the other hand, the augmented features may lie further away from the original data distribution, hence the mitigated results. 

\subsection{On the importance of the calibration set}\label{sec:oodround}
\begin{table*}[!t]
\caption{We measure the impact of the data "quality". We consider training on a subset of the test set of ImageNet (ImNet val), the standard train set (ImNet train), some adversarial \cite{hendrycks2021natural} and out-of-distribution \cite{srivastava2022out} sets for ImageNet models. Finally, we considered two extreme scenarios: Mnist \cite{deng2012mnist} (rescaled to $224 \times 224$) and white noises. We use the same quantization as in Table \ref{tab:ambiguity}.}
\label{tab:ood}
\centering
\setlength\tabcolsep{4pt}
\begin{tabular}{|c|c|c|c|c|c|c|c|}
\hline
& Architecture & ImNet (val) & ImNet (train) & ImNet (adv) & ImNet (ood) & Mnist & White Noise \\
\hline
\hline
\multirow{3}{*}{AdaRound} & ResNet 50 & 62.908 & 61.318 & 64.174 & 66.030 & 18.118 & 6.944 \\
& MobileNet v2 & 66.428 & 65.314 & 67.068 & 67.084 & 37.494 & 33.466 \\
& ViT b16 & 33.258 & 31.256 & 27.262 & 26.740 & 12.762 & 29.974 \\
\hline
\hline
\multirow{3}{*}{BrecQ} & ResNet 50 & 64.996 & 63.644 & 66.644 & 68.476 & 20.939 & 8.987 \\
& MobileNet v2 & 51.200 & 50.130 & 51.888 & 51.940 & 22.071 & 18.071 \\
& ViT b16 & 60.006 & 57.952 & 54.249 & 53.047 & 39.308 & 56.391 \\
\hline
\end{tabular}
\end{table*}
In order to evaluate the influence of the calibration distribution, we propose several sets to illustrate different levels of distribution shift. First, we use ImageNet test set to represent the most plausible in-distribution set. Second, we use data excerpted from the standard training set which corresponds to the more practical use-case. Then, we use the two natural out-of-distribution datasets introduced for ImageNet. \citet{hendrycks2021natural} introduced a set of natural examples that are adversarial examples to most convolutional neural networks and transformers to a lesser extent. Similarly, \citet{srivastava2022out} provided a set of images that are out-of-distribution with respect to the standard ImageNet training set. These two calibration sets represent realistic setups for default data points for a GPTQ method applied to computer vision models. Finally, we consider to extreme scenarios with rescaled grayscale images from MNIST \cite{deng2012mnist} and white noise images. 
In Table \ref{tab:ood}, we report our empirical results obtained with AdaRound and BrecQ. Our observations are three-fold. First, using the ImageNet test set only provides marginal benefits that is not attributed to better generalization capacities but rather to directly fitting on (a fraction of) the test examples. Second, we observe that natural images, although out-of-distribution, do not significantly degrade GPTQ performance. Conversely, we observe a similar behavior as in \citep{yang2022dataset} where optimizing on more challenging examples leads to a higher accuracy. It is important, however, to note that these examples are not challenging for transformer architectures, which explains the performance drop on ViT. Third, although MNIST and white noise lead to significant performance drops at challenging compression rates (W4/A4), we also observed during our experiments, decent performance in less challenging setups. Formally, using W4/A8 on ResNet we observe that optimization on AdaRound only leads to a 1.243 accuracy drop which places the resulting accuracy 7 points over the performance of a baseline data-free quantization. These results explain, for instance, the good performance of data generation for data-free quantization \cite{yin2020dreaming,li2021mixmix}.
Consequently, our insight regarding the choice of calibration set is as follows: \textit{GPTQ methods are robust to the distribution of the calibration set.} 

\subsection{On the importance of weight ambiguity}\label{sec:gradround}

In this section, we test the underlying hypothesis of \citep{squant2022}, which suggests that weight ambiguity is a valid indicator of the importance of a weight tensor during the quantized inference. We propose to define the level of ambiguity of a weight value $w \in W$ as the distance between the scaled weight value $\frac{w}{s_W}$ and its quantized counterpart $\left\lceil\frac{w}{s_W}\right\rfloor$. Intuitively, this distance provides an estimate of the information that gets lost from the rounding operation. Consequently, we would assume that the optimal quantization process would simply round the least ambiguous values to the nearest element and perform another modification on ambiguous values. This is the intuition behind SQuant \cite{squant2022}, a state-of-the-art data-free technique. The authors propose to change the rounding operation, only on the most ambiguous values of some layers. In order to test the relevance of this hypothesis, we propose to apply a masking to AdaRound such that only the most ambiguous weight values are optimized. We hyper-tuned the percentage of the most ambiguous values on a second calibration set and find optimal values at 0.5\% (\textit{i.e.} only focus on the 0.5\% most or least ambiguous weights)

In Table \ref{tab:ambiguity}, we report the accuracy of the GPTQ methods where we specifically focus on non-ambiguous or ambiguous weight values during the optimization process (by freezing the other category). We observe that this leads to no significant benefits in terms of accuracy nor convergence speed of the algorithm. This can be attributed to the fact that, intuitively, the optimization process will change the rounding operation on ambiguous values more often and will need to also change a few non-ambiguous weights to compensate accordingly.
\begin{table}[!t]
\caption{Performance of GPTQ methods based on their focus on weights based on the ambiguity notion. We either optimize the 0.5\% least ambiguous (non-ambiguous column) or the 0.5\% most ambiguous (ambiguous column) weight values. We use W4/A4 for ResNet 50 and ViT while we use W4/A8 for MobileNet v2.}
\label{tab:ambiguity}
\centering
\setlength\tabcolsep{4pt}
\begin{tabular}{|c|c|c|c|}
\hline
Architecture & non-ambiguous & ambiguous & baseline \\
\hline
\hline
\multicolumn{4}{|c|}{AdaRound}\\
\hline
\hline
ResNet 50 & 61.002 & 60.452 & 61.318 \\
MobileNet v2 & 64.547 & 64.712 & 65.314 \\
ViT b16 & 31.103 & 31.045 & 31.256 \\
\hline
\hline
\multicolumn{4}{|c|}{BrecQ}\\
\hline
\hline
ResNet 50 & 62.972 & 62.808 & 63.644 \\
MobileNet v2 & 49.761 & 50.002 & 50.130 \\
ViT b16 & 57.904 & 56.180 & 57.952 \\
\hline
\end{tabular}
\end{table}
Consequently, we conclude that \textit{data-free assumptions on weight ambiguity do not translate to straightforward guidelines for GPTQ improvements}. 

\subsection{On the importance of weight magnitude}\label{sec:magniround}
In a similar manner, we study the relevance of focusing the optimization on weight values based on their magnitude by applying a mask based on the weight magnitude. We calibrate the percentage of values to mask on a second calibration set and found that it is best to mask 10\% of the values. 

In Table \ref{tab:magnitude}, we report the influence of this masking on the accuracy. We observe the magnitude of the weights does not affect the GPTQ methods which can be attributed to the fact that the rounding error propagates equally regardless of the magnitude of the weight. This can be put in perspective with recent work on non-uniform quantization \cite{yvinec2023power} which demonstrate the importance of having a decent precision on low magnitude values. However, in our case, with uniform quantization, it appears that focusing on such weight values is counter productive.
\begin{table}[!t]
\caption{Performance of GPTQ methods with respect to the focus on weights by their magnitude. We either optimize the 10\% smallest values (low values) or the 10\% highest values (high values). We use the same quantization as in Table \ref{tab:ambiguity}.}
\label{tab:magnitude}
\centering
\setlength\tabcolsep{1.5pt}
\begin{tabular}{|c|c|c|c|c|}
\hline
 & Architecture & low values & high values & baseline \\
\hline
\hline
\multirow{3}{*}{AdaRound} &ResNet 50 & 60.402 & 59.481 & 61.318 \\
&MobileNet v2 & 64.836 & 64.119 & 65.314 \\
&ViT b16 & 30.614 & 30.166 & 31.256 \\
\hline
\hline
\multirow{3}{*}{BrecQ} &ResNet 50 & 63.216 & 62.515 & 63.644 \\
&MobileNet v2 & 50.077 & 49.653 & 50.130 \\
&ViT b16 & 57.090 & 56.991 & 57.952 \\
\hline
\end{tabular}
\end{table}

We conclude that, contrary to insights from non-uniform quantization, it appears that \textit{weight magnitude is not an effective, straightforward indicator for GPTQ improvements}.

\subsection{On the importance of the bias}\label{sec:biasround}
In post-training quantization, it is commonly admitted that the biases play a crucial role in the performance of the quantized model. Bias correction, introduced by Nagel \textit{et al.} \cite{nagel2019data}, edits the bias values in order to preserve the mean (per neuron) of each layer outputs by a simple update. We propose to investigate the benefits of optimizing the weights and biases jointly in GPTQ. 
We define the constraint over the bias $b$ as the range of values that can be explored. We learn a parameter $\epsilon_b \in ]-\alpha |b| ; \alpha |b|[$. Consequently, for $\alpha = 0$ we do not optimize the biases which corresponds to the baseline approach. In Table \ref{tab:bias}, we report the performance for several values of $\alpha$ and observe unilateral results. Biases should not be learned: we know that deep neural networks are particularly good at extracting biases from data \cite{shah2020pitfalls}. As we use a very small calibration set, the optimization defeats the generalization capacities of the models. 
\begin{table}[!t]
\caption{Performance of GPTQ methods with different constraint on the optimization of the biases. For a constraint $\alpha = 0$, we do not update the biases which corresponds to the baseline methods.}
\label{tab:bias}
\centering
\setlength\tabcolsep{4pt}
\begin{tabular}{|c|c|c|c|c|}
\hline
Architecture & baseline & $\alpha = 0.33$ & $\alpha = 0.66$ & $\alpha = 1$ \\
\hline
\hline
\multicolumn{5}{|c|}{AdaRound} \\
\hline
\hline
ResNet 50    & 61.318 & 30.423 & 0.100 & 0.100 \\
MobileNet v2 & 65.314 & 11.612 & 0.100 & 0.100 \\
ViT  b16        & 31.256 & 9.742 & 0.100 & 0.100 \\
\hline
\hline
\multicolumn{5}{|c|}{BrecQ}\\
\hline
\hline
ResNet 50    & 63.849 & 26.836 & 0.100 & 0.100 \\
MobileNet v2 & 50.130 & 5.010  & 0.100 & 0.100 \\
ViT b16         & 57.952 & 7.930  & 0.100 & 0.100 \\
\hline
\end{tabular}
\end{table}
As a result, we assert that \textit{GPTQ methods should avoid the optimization of biases and stick to the straightforward bias correction as already performed in the baseline}.

\subsection{On the importance of the optimizer choice}\label{sec:optround}
\begin{table*}[!t]
\caption{Performance of GPTQ methods with different gradient-based optimizers (we recall that the default option Adam). For all the considered setups, we use the default parameters in order to avoid hyper-tuning and unfair comparisons. We use the same quantization as in Table \ref{tab:ambiguity}. We highlight in bold the results that show an improvement over the baseline (Adam).}
\label{tab:optimizer}
\centering
\setlength\tabcolsep{4pt}
\begin{tabular}{|c|c|c|c|c|c|c|c|c|c|}
\hline
& Architecture & SGD & Nesterov & Adam & AdamW & Adamax & AdaGrad & AdaDelta & RMSProp \\
\hline
\hline
\multirow{3}{*}{AdaRound} & ResNet 50    & 60.616 & 60.384 & 61.318 & \textbf{61.434} & \textbf{62.830} & \textbf{62.102} & 49.984 & 61.924 \\
& MobileNet v2 & 38.430 & 31.962 & 65.314 & \textbf{65.374} & \textbf{65.824} & 14.232 & 1.304  & \textbf{66.236} \\
& ViT b16         & 8.326  & 8.530  & 31.256 & 30.462 & \textbf{39.234} & 9.916  & 4.654  & 30.328 \\
\hline
\hline
\multirow{3}{*}{BrecQ} & ResNet 50    & 62.653 & 62.428 & 63.644 & \textbf{63.849} & \textbf{65.374} & \textbf{64.677} & 53.293 & 63.625 \\
& MobileNet v2 & 22.807 & 17.014 & 50.130 & \textbf{50.279} & \textbf{50.616} & 0.100 & 0.100 & \textbf{50.184} \\
& ViT b16         & 34.849 & 35.516 & 57.952 & 56.804 & \textbf{59.325} & 36.627 & 31.517 & 56.274 \\
\hline
\end{tabular}
\end{table*}
The optimizer selected for DNNs training have played a significant role in the increased performance of trained models. Older methods such as SGD \citep{ruder2016overview}, Nesterov \cite{sutskever2013importance}, AdaGrad \cite{duchi2011adaptive} AdaDelta \cite{zeiler2012adadelta} and RMSProp \cite{rmsprop} were prone to unstable results which lead to the wide adoption of Adam \cite{kingma2014adam}. Recent iterations have improved Adam over several aspects, such as AdamW \cite{loshchilov2017decoupled} which has been leveraged in many transformer papers \cite{touvron2021going}. However it is still unclear which optimizer should be selected in the general case.

In Table \ref{tab:optimizer}, we observe that the AdaMax optimizer \cite{kingma2014adam} systematically outperforms the original Adam and the other benchmarked optimizers. This optimizer was introduced simultaneously to Adam, and instead of normalizing the gradients with respect to the $L^2$ norm of the past values, AdaMax normalizes with respect to the infinite norm. Theoretically, this increases the robustness to small perturbations of the gradients. In GPTQ, this noise can be attributed to the quantization process itself. As a result, AdaMax is the more appropriate choice for GPTQ methods regardless of the architecture.

In conclusion, \textit{our results highlights both the importance of the optimizer choice and the systematic added value from AdaMax}. Considering the impact played by the optimization of the weight values on uniform quantization and the fact that non-uniform quantization baselines are known for significantly higher performance, in what follows, we propose a solution to generalize GPTQ to all quantization formats.

\begin{table*}[!t]
\caption{optimization of $\epsilon$ in $[0;1]$ \textit{v.s.} $\mathbb{R}$ with non-uniform quantization such as logarithmic quantization (log) low-bit floating point representation (4 bits float) and power quantization \cite{yvinec2023power}. We propose to also apply the adaptation to uniform quantization as well for its memory footprint benefits. We use the same quantization as in Table \ref{tab:ambiguity}.}
\label{tab:non_uniform}
\centering
\setlength\tabcolsep{6pt}
\begin{tabular}{|c|c|c|c|c|c|c|c|c|}
\hline
& \multicolumn{4}{c|}{$\epsilon \in [0;1]$}& \multicolumn{4}{c|}{$\epsilon \in \mathbb{R}$}\\
\hline
Architecture & uniform & log & float & power & uniform & log & float & power \\
\hline
\hline
ResNet 50 & 61.318 & 0.304 & 45.758 & 74.892 & 61.396 & 60.688 & 60.170 & 68.546 \\
MobileNet v2 & 65.314 & 0.100 & 22.134 & 0.788 & 64.790 & 64.866 & 65.028 & 66.014 \\
ViT b16 & 31.256 & 0.138 & 6.056 & 77.214 & 33.018 & 35.364 & 32.082 & 79.578 \\
\hline
\end{tabular}
\end{table*}
\subsection{On the importance of bounding the optimization to $[0;1]$}\label{sec:dsqround}

When we optimize $\epsilon \in[0;1]$, we make the implicit assumption that a slight change of $\epsilon$ leads to the same information change regardless of the magnitude of the corresponding weight scalar value $w$. While this is true in uniform quantization, this is no longer the case when we try to combine GPTQ with non-uniform quantization. In NUPES \cite{yvinec2023nupes}, authors, proposed a solution for power quantization: optimize $\epsilon \in \mathbb{R}$ with a soft approximation for the rounding process \cite{gong2019differentiable} with a sttepness defined by parameter $\beta$. Intuitively, a slight change in the value of $\epsilon$ may lead to a significant modification on the predictive function, for instance if the corresponding weight value was large.

Although, it was designed for power quantization, optimizing $\epsilon in \mathbb{R}$ still achieves good performance and finds different quantized weight values with all other representations.
We provide a discussion on the intuition behind this result in Appendix A.
In Table \ref{tab:non_uniform}, we observe the necessity to allow for more flexibility  on the values of $\epsilon$ for non-uniform quantization. For instance, logarithmic quantization with $\epsilon \in[0;1]$ achieves near random performance while with $\epsilon in \mathbb{R}$ performs on par with uniform $\epsilon \in[0;1]$ and even outperforms it on transformers. Similarly, low bit floating point also benefit from $\epsilon in \mathbb{R}$. Their initial better performance comes from the fact that their distribution is more similar to uniform quantization: low bit floating points are piece-wise uniform. We use $\beta = 20$ for power quantization, $\beta = 30$ for floating points and $\beta = 50$ for the log. A key element, to note here, is the fact that the less uniform the distribution (e.g. log) the higher the value of the steepness should be. In practice, the best value of $\beta$ varies between $20$ and $50$ but using a default value of $50$ is sufficient to reach near optimal results.

In summary, \textit{the optimization of $\epsilon \in \mathbb{R}$ for GPTQ methods enables us to leverage non-uniform quantization in order to further improve GPTQ methods}.

\begin{table}[!t]
\caption{Evaluation of the proposed mixed-precision method.}
\label{tab:mp}
\centering
\setlength\tabcolsep{4pt}
\begin{tabular}{|c|c|c|c|c|}
\hline
& \multicolumn{2}{c|}{AdaRound} & \multicolumn{2}{c|}{BrecQ} \\
\hline
Architecture & fixed & mixed & fixed & mixed \\
\hline
\hline
ResNet 50    & 61.318 & \textbf{64.675} & 63.849 & \textbf{66.181} \\
MobileNet v2 & 65.314 & \textbf{67.402} & 50.130 & \textbf{56.756} \\
ViT b16         & 31.256 & \textbf{35.179} & 57.952 & \textbf{59.000} \\
\hline
\end{tabular}
\end{table}
\begin{table*}[!t]
\caption{Summary of the added value from the best practices we propose. All results are averaged over 10 runs and all standard deviations are below $0.5$\% accuracy}
\label{tab:summary}
\centering
\setlength\tabcolsep{4pt}
\begin{tabular}{|c|c|c|c|c|}
\hline
& AdaRound & BrecQ & NUPES & fp32 \\
\hline
\hline
ResNet 50 (W4/A4)   & 61.318  $\rightarrow$ 65.107  (\textbf{+3.789}) & 63.644 $\rightarrow$ 66.253  (\textbf{+2.609}) & 68.546 $\rightarrow$ 70.298  (\textbf{+1.752}) & 76.150 \\
MobileNet v2 (W4/A8) & 65.314 $\rightarrow$ 68.380  (\textbf{+3.066}) & 50.130 $\rightarrow$ 56.756  (\textbf{+6.626}) & 66.014 $\rightarrow$ 67.344 (\textbf{+1.330}) & 72.074 \\
ViT b16 (W4/A4)     & 31.256 $\rightarrow$ 38.075  (\textbf{+6.819}) & 57.952 $\rightarrow$ 59.895  (\textbf{+1.943}) & 79.578 $\rightarrow$ 80.201  (\textbf{+0.623}) & 80.978 \\
\hline
\end{tabular}
\end{table*}

\subsection{On the importance of mixed-precision}\label{sec:mpround}
Mixed precision consists in the search for the adequate bit width for each operation. Its granularity can go from the single neuron to the computational block. Previous work \citep{wang2019haq}, have shown its efficiency in data-free and QAT set-ups, however its integration within the GPTQ framework remains to be addressed. 
We propose a novel importance-based mixed-precision technique that works with GPTQ methods. For each neuron $n_i$, we measure compute the gradients $g_i$ with respect to the model outputs as an estimation of the sensitivity of the model predictions with respect to a given neuron. This stems on the intuition from attribution \cite{selvaraju2016grad} on neural networks inference and its compression application \cite{yvinec2022singe}. Based on this estimation, we need to derive bit-widths for each neuron $n_i$.

We propose to leverage first and second order statistics, \textit{i.e.} the mean $\mu$ and standard deviation $\sigma$. The final precision as a number of bits $b_i$ is given by the target average number of bits $b$ and the distance to the mean in terms of standard deviations, \textit{i.e.} $b_i = b + \left\lfloor\frac{g_i - \mu}{\sigma}\right\rceil$ where $\lfloor\cdot \rceil$ rounds towards zero (e.g. $\lfloor 0.9 \rceil = 0$, $\lfloor -0.9 \rceil = 0$ or $\lfloor 2.1 \rceil = 2$). 
Intuitively, if we have two neurons $n_i$ and $n_j$ such that $g_i = \mu$ and $g_j = \mu + \sigma$ then neuron $n_j$ is significantly more sensitive than neuron $n_i$. For this reason, we increase the quantization precision accordingly.
We then normalize the bit widths $b_i$ such that the average exactly matches the target bit-width $b$. 

As shown in Table \ref{tab:mp}, the proposed mixed-precision significantly improves the performance of GPTQ methods. In particular, on ResNet 50 quantized in 4 bits on average, the mixed-precision enables AdaRound to outperform the fixed-precision BrecQ which indicates that mixed-precision on its own outperforms the added value of BrecQ on models for which both methods are relevant. All this was obtained at a fraction of the computational cost of the whole method: mixed-precision increase the processing time by 3.2\% on average.

\begin{figure}
    \centering
    \includegraphics[width = \linewidth]{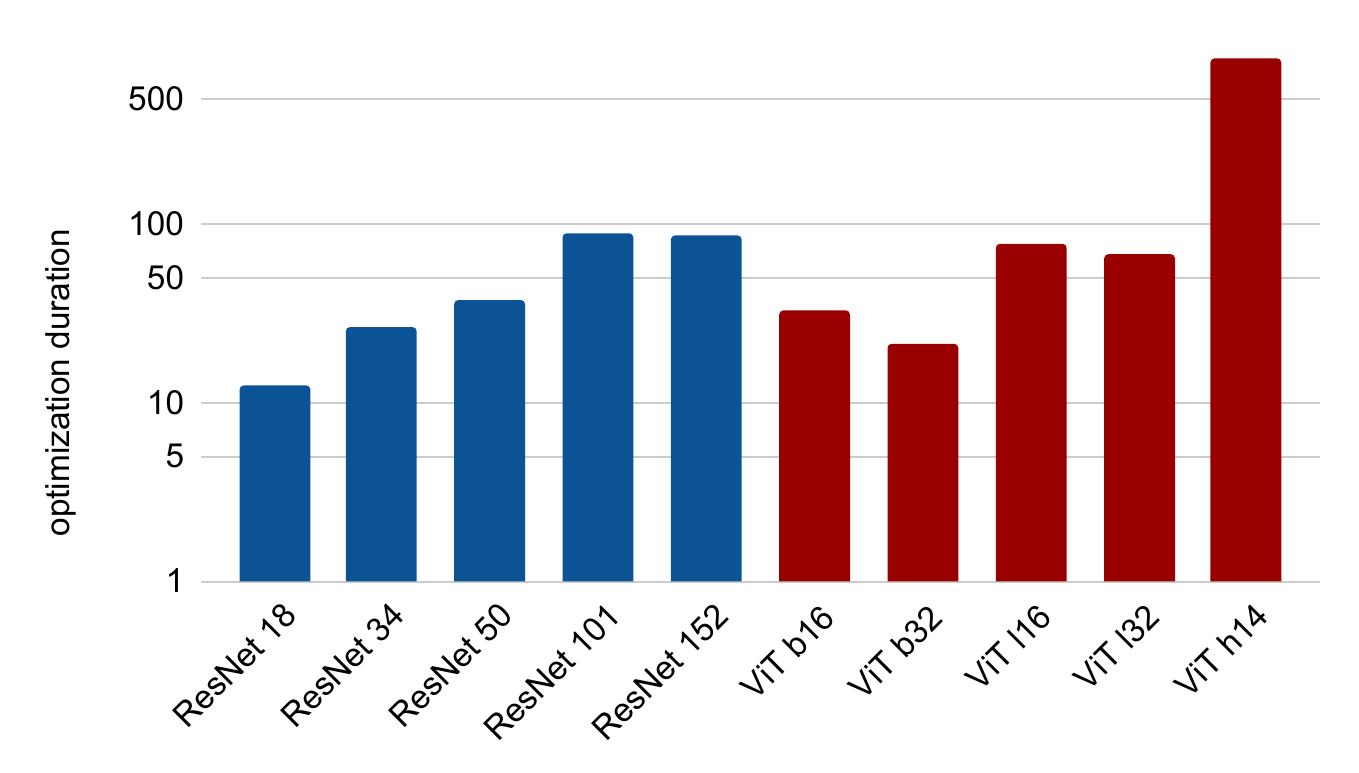}
    \caption{Optimization duration (in minutes, using 10K steps) with every elements directly loaded a single A100 gpu.}
    \label{fig:duration}
\end{figure}

\section{Putting it together}

We summarize the benefits from using the best practices introduced in this paper, namely: no weight selection (neither on weight ambiguity or magnitude), no feature augmentation, using the standard calibration set, $l_2$ loss, no bias optimization, Adamax optimizer, optimizing on $\epsilon \in \mathbb{R}$, gradient-based mixed precision. Results are showcased on Table \ref{tab:summary}. Using the provided guidelines allows to significantly improve the performance of AdaRound, BrecQ and and the recent NUPES on every architecture. Using NUPES, for instance, we are able to quantize ViT in W4/A4 losing less than $1\%$ in accuracy, which is a massive improvement over previous state-of-the-art. Thus, the ideas and best practices presented in this paper shall pave the way to design more efficient, and scalable, GPTQ methods. In what follows, we provide concluding remarks and discuss reflections and limitations of current GPTQ techniques. 

\section{Discussion and conclusion}

\paragraph{conclusion:} in this study, we identified GPTQ as a suitable trade-off between data-free quantization techniques and more accurate, yet expensive QAT methods. We challenged several methodological choices made in recent GPTQ papers. In particular, we showed GPTQ is essentially robust to weight selection, feature augmentation as well as the choice of calibration set. Moreover, we also derived a number of best practices to improve current and future methods, regarding the problem formulation (loss, degrees of freedom, use of non-uniform quantization schemes) or optimization (choice of variable and optimizer). Lastly, we proposed a novel importance-based mixed-precision technique. Following these best practices, we were able to significantly improve the performance of several GPTQ methods on a number of deep neural network architectures, paving the way for research on efficient quantization of large networks, e.g. LLMs. However, they bear a number of limitations.

\paragraph{Limitations and perspectives:} the first limitation for GPTQ methods is the scalability w.r.t. model size (from e.g. 25M parameters ResNet 50 to 600M parameters ViT h14). In Figure \ref{fig:duration}, we report the optimization duration for several architectures. First, our measurements are performed with every data already loaded on the gpu which speeds up the process but is not feasible with some models such as LLMs and diffusion models. Second, we observe that the duration can already tower over 10 hours for ViT h14. Consequently, we argue that GPTQ methods should not aim at increasing the optimization duration but rather, further reduce it.

A second limitation is the focus on weight values only. GPTQ methods learn how to rounding up or down the weight values but never change the quantization rule of the activations. This is a significant limitation as the quantization of the activation, that can lead to massive accuracy drops. In particular, some implementations of AdaRound and BrecQ showcase very different performance due to the fact that some only quantize the outputs of each layer leading to higher bit-width on the inputs of the subsequent layers from the skip connections. More specifically, if the first layer is quantized in W8/A8 and the next layers in W4/A4, if we do not quantize the inputs but only the outputs, then all layers connected to the first one will actually perform in W4/A8 due to the skip connections. This can lead to more than 10 points difference in accuracy.
On the other hand, it is known that LLMs are very sensitive to activations quantization due to the presence of outliers \citep{dettmers2022llm}. So much so, that quantization techniques that tackle LLMs quantization do not manage to perform proper int8 quantization of the activations. Similarly, EfficientNets are also sensitive to activation quantization \citep{yvinec2022spiq}. Consequently, we argue that activation quantization will play a crucial role in the future of GPTQ methods. However, changing the rounding operation of the activation may have a negative impact on the latency, in Appendix B, we detail how this could be efficiently implemented. 

Third, GPTQ methods are performed layer per layer from the first to the last layer of the model. The intuition behind this implementation is the fact that once the first layers are set, the quantized inputs correspond exactly to the intermediate features that will be computed at inference. In other words, when we optimize a layer, its quantized inputs are exactly what it would receive at test time. Formally, GPTQ methods are designed for directed acyclic graphs (DAGs). However, many trending models are not DAGs such as diffusion models \citep{schuhmann2021laion} and LLMs. As a result, some assumptions made in GPTQ methods do not hold. We hypothesize that adapting GPTQ to non-DAGs will lead to significant performance improvements in the future.

\bibliography{aaai24}

\newpage
~
\newpage

\section*{A. Intuition behind the use of QAT in QPTQ}\label{sec:intuition}
Let $W\in\R{n\times m}$ be a weight tensor, $X\in\R{m}$ an input tensor and $\sigma$ the corresponding activation function. The quantized inference is simulated as
\begin{equation}\label{eq:quantization_base}
    {\left(\round{\frac{W}{s_W}} \times \round{\frac{X}{s_X}}\right)}_i = \sum_j {\left(\round{\frac{W}{s_W}}\right)}_{i,j} \times {\left(\round{\frac{X}{s_X}}\right)}_{j}
\end{equation}
For the sake of simplicity, we assume that the scales are all equal to $1$. Then, equation \ref{eq:quantization_base} becomes
\begin{equation}\label{eq:quantization_simplified}
    {\left(\round{W} \times \round{X}\right)}_i = \sum_j {\left(\round{W}\right)}_{i,j} \times {\left(\round{X}\right)}_{j}
\end{equation}
In AdaRound \cite{nagel2020up}, the optimization minimizes the quantization error at the activation level, \textit{i.e.} we minimize
\begin{equation}\label{eq:adaround_goal}
    \mathcal{L}_{\text{adaround}}(W,X) = \left\| \sigma\left(W\times X\right) - \sigma\left(\round{W} \times \round{X}\right) \right\|
\end{equation}
Let's denote the quantization errors $\epsilon_W$ and $\epsilon_X$ such that $W = \floor{W} + \epsilon_W$ and $X = \floor{X} + \epsilon_X$ respectively. The difference $\epsilon_{W\times X}$ between $W\times X$ and its quantized counterpart $\round{W} \times \round{X}$ can be decomposed in three terms
\begin{equation}\label{eq:quantization_error}
    \epsilon_{W\times X} = \epsilon_W \times \floor{X} + \floor{W} \times \epsilon_X + \epsilon_W \times \epsilon_X.
\end{equation}
The error diminishes, through the optimization defined in equation \ref{eq:adaround_goal}, as we update $\epsilon_W$. In post-training quantization, following \cite{nagel2020up}, it is a common practice to optimize $\mathcal{L}_{\text{adaround}}$ under the constraint $\epsilon_{W} \in [0;1]^{n\times m}$. This search space allows to systemically find an optimal solution to equation \ref{eq:adaround_goal} if and only if $n=m=1$. Otherwise, we can build simple examples that require negative values for $\epsilon_W$ to be optimal. Let's consider $n=1$ and $m=2$, then
\begin{equation}
    \begin{pmatrix}
    4.1 & 3.2
    \end{pmatrix}\times
    \begin{pmatrix}
    6.4\\2.4
    \end{pmatrix} = 33.92
\end{equation}
is best approximated by the quantized weight matrix $\begin{pmatrix} 5 & 2\end{pmatrix}$ in the following operation
\begin{equation}
    \begin{pmatrix}
    5 & 2
    \end{pmatrix}\times
    \begin{pmatrix}
    6\\2
    \end{pmatrix} = 34.
\end{equation}
This is the only solution as we do not change the quantization of the inputs (for inference speed performance). It requires $\epsilon_W = \begin{pmatrix} 1 & -1\end{pmatrix} \notin [0;1]^{n\times m}$.

\section*{B. Up or Down? Activation quantization}\label{sec:adaround_activation}
Let's consider a layer defined by the matrix multiplication $W\times X$. Its quantized counterpart computes $Q^{-1}(\hat W\times \hat X)$ where $\hat W$ and $\hat X$ are quantized. In practice, quantization research implements this operation as follows
\begin{equation}
    Q^{-1}(\hat W\times \hat X) = (s_W s_X) \times \hat W\times \left\lfloor \frac{X}{s_X}\right\rceil
\end{equation}
where $s_W$ and $s_X$ are scaling factors encoded in floating point with high precision (16 or 32 bits). This is a simulation of the actual behavior at inference which does not include any operations in floating point. In the inference engine, the scaling term $s_W s_X$ is converted into an integer scaling followed by a bitshift. As a result the next layer gets as an input a tensor $X$ which is already quantized. Consequently, we get 
\begin{equation}
    Q^{-1}(\hat W\times \hat X) = 2^{-e} M \times \hat W\times \hat X
\end{equation}
Consequently, in order to encode a round-up \textit{versus} round-down on the activations, we propose to apply an additive mask $\mathcal{M}$ with binary values, which leads to the following forward pass
\begin{equation}
    Q^{-1}(\hat W\times \hat X) = 2^{-e} M \times \hat W\times \hat X + \mathcal{M}
\end{equation}
This is equivalent to the addition of a bias term with the same shape as the inputs.
As a result, the overhead is equal to the activation sizes quantized in binary values.

\end{document}